\documentclass[letterpaper]{article} 
\usepackage[preprint]{aaai2027}  
\usepackage[hyphens]{url}  
\usepackage{graphicx} 
\usepackage{natbib}  
\usepackage{caption} 
\usepackage{microtype}
\usepackage{subcaption}
\usepackage{booktabs} 
\usepackage{multirow}

\usepackage{algorithm}
\usepackage{algpseudocode}

\usepackage{amsmath}
\usepackage{amssymb}
\usepackage{mathtools}
\usepackage{amsthm}

\usepackage[capitalize,noabbrev]{cleveref}

\theoremstyle{plain}

\theoremstyle{definition}

\theoremstyle{remark}

\usepackage{float}
\usepackage{tcolorbox}
\usepackage{booktabs}
\usepackage{multirow}
\usepackage{siunitx}
\usepackage{xcolor}         
\usepackage{threeparttable} 

\usepackage{array}   
\newcolumntype{C}[1]{>{\centering\arraybackslash}p{#1}}

\usepackage{lineno}

\title{A Comparative analysis of Layer-wise Representational  \\ Capacity in AR and Diffusion LLMs}

\author{
Raghavv Goel* \quad
Risheek Garrepalli* \quad
Sudhanshu Agrawal \quad
Chris Lott \quad
Mingu Lee \quad
Fatih Porikli
}
\affiliations{
Qualcomm AI Research, Qualcomm Technologies Inc.\\
\texttt{raghgoel,rgarrepa,mingul@qti.qualcomm.com}
}
\begin{document}

\maketitle

\begin{abstract}
 Autoregressive (AR) language models build representations incrementally via left‑to‑right prediction, while diffusion language models (dLLMs) are trained through full‑sequence denoising. Although recent dLLMs match AR performance, whether diffusion objectives fundamentally reshape internal representations remains unclear. We perform the first layer‑ and token‑wise representational analysis comparing native dLLMs (LLaDA), native AR models (Qwen2.5), and AR‑initialized dLLMs (Dream‑7B), using cosine similarity across layers and tokens alongside static inference-time layer-skipping as an analytical probe of redundancy. We find that diffusion objectives produce more global representations with substantial early‑layer redundancy and reduced recency bias, while AR objectives yield tightly coupled, locally-structured representations. AR‑initialized dLLMs retain AR‑like dynamics despite diffusion training, revealing persistent initialization bias. Leveraging this redundancy, native dLLMs absorb up to 18.75\% FLOPs reduction: retaining over 90\% performance on math-reasoning and coding benchmarks, while AR models collapse under identical skipping, revealing that diffusion objectives, rather than architecture alone, induce depth redundancy that enables principled compression
\end{abstract}

\section{Introduction}
\noindent\textbf{From Next-Token Prediction to Diffusion Objectives.}
Autoregressive (AR) language models are trained via next-token prediction (NTP), constructing representations incrementally through left-to-right factorization. In contrast, diffusion language models (dLLMs) replace this causal factorization with full-sequence denoising, iteratively refining an entire token sequence from noise to data. Recent discrete diffusion models—such as LLaDA~\cite{Nie2025LLaDA} and DiffuCoder~\cite{gong2025diffucoder}—have demonstrated performance on par with strong AR baselines across a range of downstream tasks. Despite this progress, a fundamental question remains unanswered: does training with a diffusion objective materially change how language models organize and abstract information internally, or are diffusion models representationally similar to AR models once performance is matched?

\noindent\textbf{The representational gap.}
Most prior work on diffusion LLMs has focused on efficiency advantages—parallel decoding, verifier-based sampling, or architectural optimizations—rather than the structure of their learned representations. As a result, we lack a systematic understanding of how diffusion objectives shape internal geometry across depth and tokens, and how this differs from AR training. While recent studies~\cite{gong2025diffucoder} have begun to explore local versus global behaviors in diffusion models, a comprehensive, layer- and token-wise comparison between AR and diffusion objectives is still missing. Such an analysis is crucial: representational structure reflects not only inference behavior but also latent inductive biases introduced by training objectives, much like how residual connections or normalization schemes influence optimization and generalization in deep networks.

\noindent\textbf{Global vs.\ local representations.}
We hypothesize that the training objective itself—and in particular, whether supervision is delivered through causal next-token prediction or full-sequence denoising—fundamentally shapes representational abstraction in LLMs. We use the term \textit{global representations} to denote hidden states that integrate information from the full input sequence, rather than being dominated by the most recent tokens. Full-sequence feedback in diffusion training naturally encourages such global integration, whereas AR training enforces a strong recency bias due to its causal structure. In particular, full-sequence denoising may encourage earlier consolidation of global abstractions, concentrating representational redundancy in shallower layers—consistent with recent findings that intermediate layers can encode richer representations than final layers~\cite{skean2025layer}. To operationalize this distinction, we analyze representational similarity across layers and tokens, measuring how hidden states vary with token position and depth.

\noindent\textbf{Isolating objective from initialization.}
A central challenge in comparing AR and diffusion models is disentangling the effects of training objective from architecture and initialization. To address this, we study three families: (i)a native AR LLM trained exclusively with NTP (Qwen2.5\citep{Qwen2024Qwen25}), (ii)a native dLLM trained from scratch with a diffusion objective (LLaDA\cite{Nie2025LLaDA}), and (iii)an AR-initialized dLLM that undergoes diffusion training after AR pretraining (Dream-7B\citep{dream2025}, initialized from Qwen2.5). This design allows us to isolate whether observed representational properties arise from the diffusion objective itself or are inherited from AR initialization.

\noindent\textbf{Layer skipping as a diagnostic probe.}
Beyond static analysis, we introduce inference-time layer skipping as a controlled intervention to probe representational redundancy. Importantly, we do not propose layer skipping as a deployment strategy; rather, we use it as an analytical tool to stress-test how much computation can be removed before representations and performance degrade. If early layers are redundant or globally integrated, their removal should incur limited performance loss; conversely, models with tightly coupled, locally structured representations should fail under the same perturbation. This static, task-agnostic policy requires no KV-cache sharing and no architectural modifications, providing an orthogonal complement to cache-centric methods such as YOCO~\cite{sun2024you}.

\begin{figure*}[t]
  \centering
  \begin{minipage}[t]{0.48\textwidth}
    \centering
    \includegraphics[width=\linewidth]{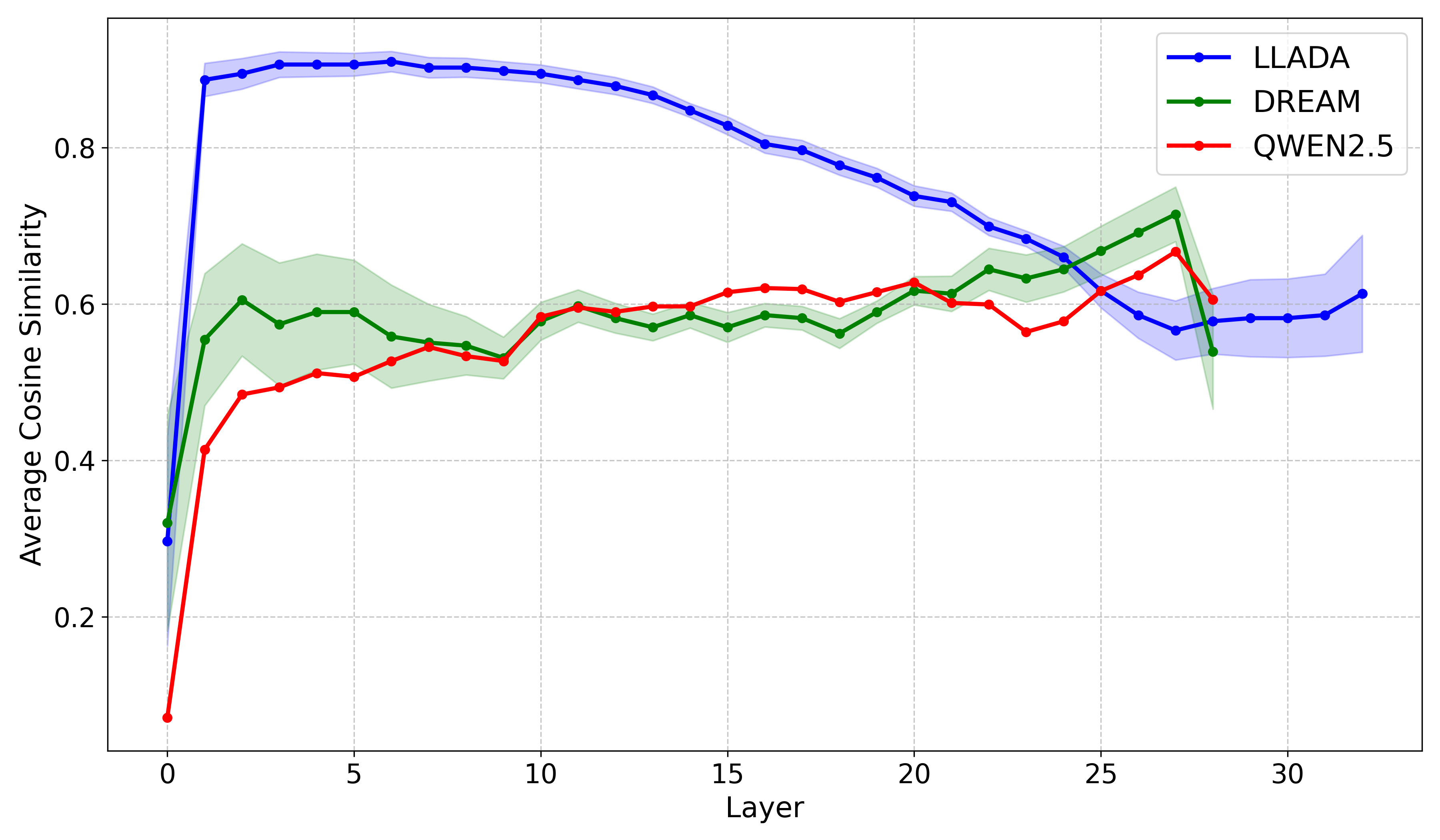}
    \vspace{-14pt}
    \subcaption{\textbf{Why dLLMs are skip-tolerant.} LLaDA (native dLLM) exhibits high cosine similarity ($>$0.9) in early layers—a redundancy plateau absent in Qwen2.5 and Dream-7B, which closely track each other throughout depth, revealing persistent AR initialization bias. Shaded regions show std.\ across denoising steps.}
    \label{fig:avg_token_similarity}
  \end{minipage}
  \hfill
  \begin{minipage}[t]{0.48\textwidth}
    \centering
    \includegraphics[width=\linewidth]{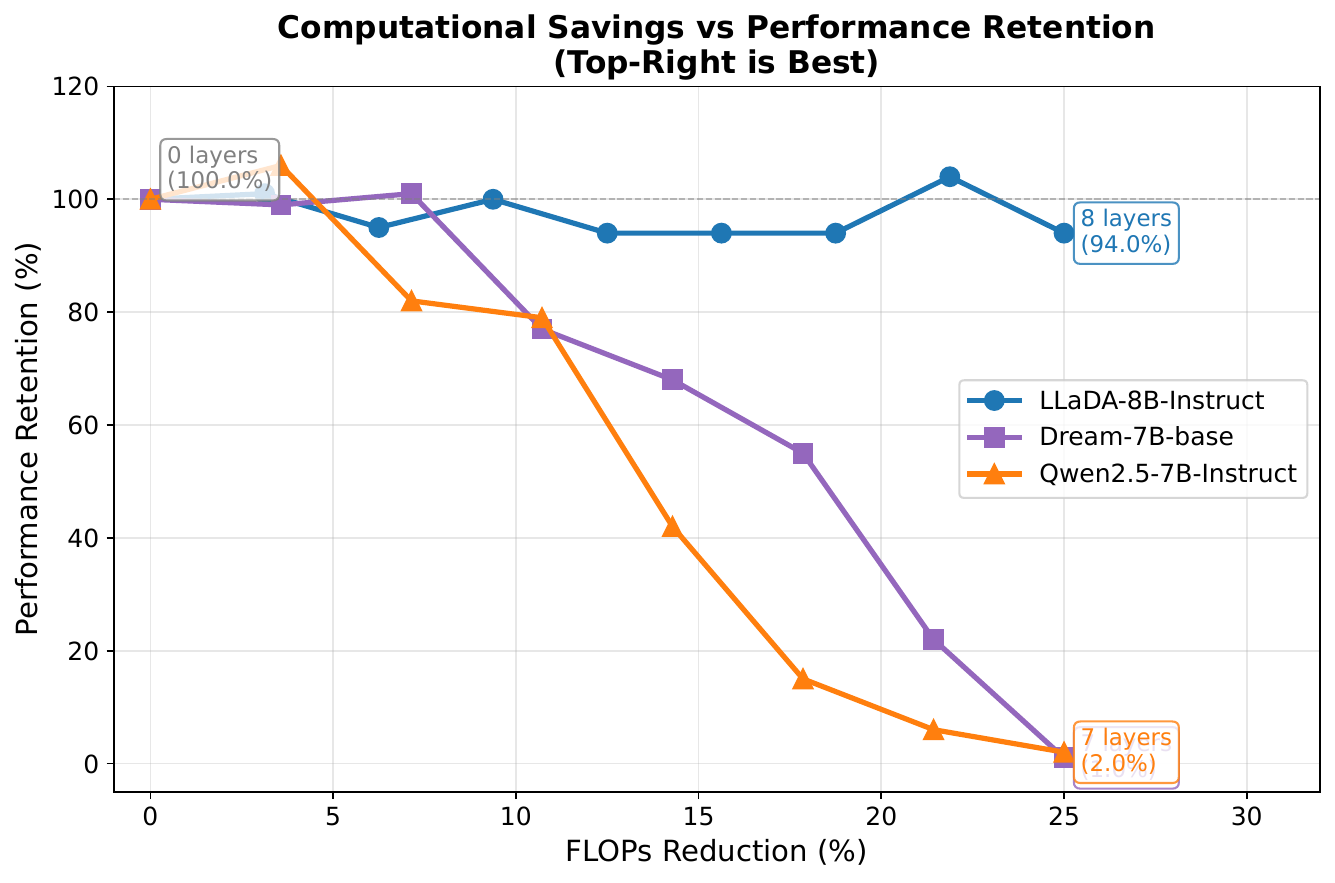}
    \vspace{-14pt}
    \subcaption{\textbf{Redundancy translates to efficiency.} LLaDA retains 94\% GSM8K performance at 18.75\% FLOPs reduction (6 layers skipped), while Qwen2.5-7B-Instruct---despite retaining 82\% at 2 layers (7.14\% reduction)---collapses to 42\% by just 4 layers (14.29\% reduction). Dream-7B-base exhibits intermediate behaviour, retaining 77--101\% at low skip rates before degrading at higher compression.}    
    \label{fig:layer-skip-flops}
  \end{minipage}
  \caption{\textbf{Representational redundancy in native dLLMs enables efficient inference-time layer skipping.}
  \textit{Left:} Layer-wise cosine similarity reveals that LLaDA develops a high-similarity plateau in early layers—objective-induced redundancy absent in AR models (Qwen2.5) and AR-initialized dLLMs (Dream-7B).
  \textit{Right:} This structural redundancy directly enables skip tolerance: native dLLMs absorb aggressive layer skipping with minimal quality loss, while AR models degrade sharply under identical conditions.}
  \label{fig:combined}
\end{figure*}

\noindent\textbf{Our findings.}
Across all analyses, a consistent picture emerges. Native dLLMs learn globally coherent representations with substantial early-layer redundancy, exhibiting reduced recency bias and high cross-token similarity in early layers. AR-initialized dLLMs, by contrast, retain AR-like representational dynamics even after diffusion training, demonstrating a persistent imprint of initialization and confirming that global redundancy is not a trivial consequence of architecture alone. These representational differences translate directly into measurable efficiency gaps: native dLLMs tolerate up to 18.75\% FLOPs reduction via layer skipping while retaining over 90\% performance, whereas AR models collapse under identical interventions.

\noindent\textbf{Contributions.} We summarize our contributions as follows:

\begin{itemize}
\item \textit{Representational analysis revealing objective-induced redundancy.} We present the first systematic layer-wise and token-wise similarity analysis comparing native dLLMs, AR models, and AR-initialized dLLMs. We show that diffusion objectives produce more global representations with concentrated early-layer redundancy and minimal recency bias, while AR objectives maintain incremental, locally structured refinement with strong recency bias throughout depth. We further reveal a strong \textbf{initialization bias}: AR-initialized dLLMs (Dream-7B) retain AR-like representational patterns despite diffusion training, aligning more closely with Qwen2.5 than with LLaDA.

\item \textit{Inference-time layer skipping.} Leveraging objective-induced representational redundancy, we introduce a static, task-agnostic layer-skip policy requiring no KV-cache sharing and no architectural modifications. Native dLLMs (LLaDA) achieve up to 18.75\% FLOPs reduction with $<$10\% average accuracy degradation, while AR models show substantial brittleness under the same intervention, providing an architecture-agnostic complement to cache-centric designs.

\item \textit{Cross-domain benchmarking.} We evaluate across math reasoning (GSM8K, MATH-500) and code synthesis (HumanEval, MBPP), demonstrating consistent patterns: native dLLMs tolerate aggressive layer skipping (6 layers, $>$90\% average retention), AR-initialized dLLMs exhibit intermediate robustness ($\sim$78\% average retention at 2-layer skip, declining at higher compression), and native AR models degrade substantially at 2-layer skip ($\sim$63\% average retention on tasks with non-trivial baselines).
\end{itemize}

\section{Layer-wise and Token-wise Similarity Analysis}



\textbf{Motivation:} To understand how training objectives shape internal representations and induce redundancy patterns, we examine layer-to-layer similarity across dLLMs and AR models. Unlike AR models that build representations incrementally through left-to-right token prediction, dLLMs receive full-sequence gradient feedback during training, potentially leading to different abstraction pathways and redundancy structures across depth. 

\begin{figure*}[t]
  \centering
  \includegraphics[width=0.8\textwidth]{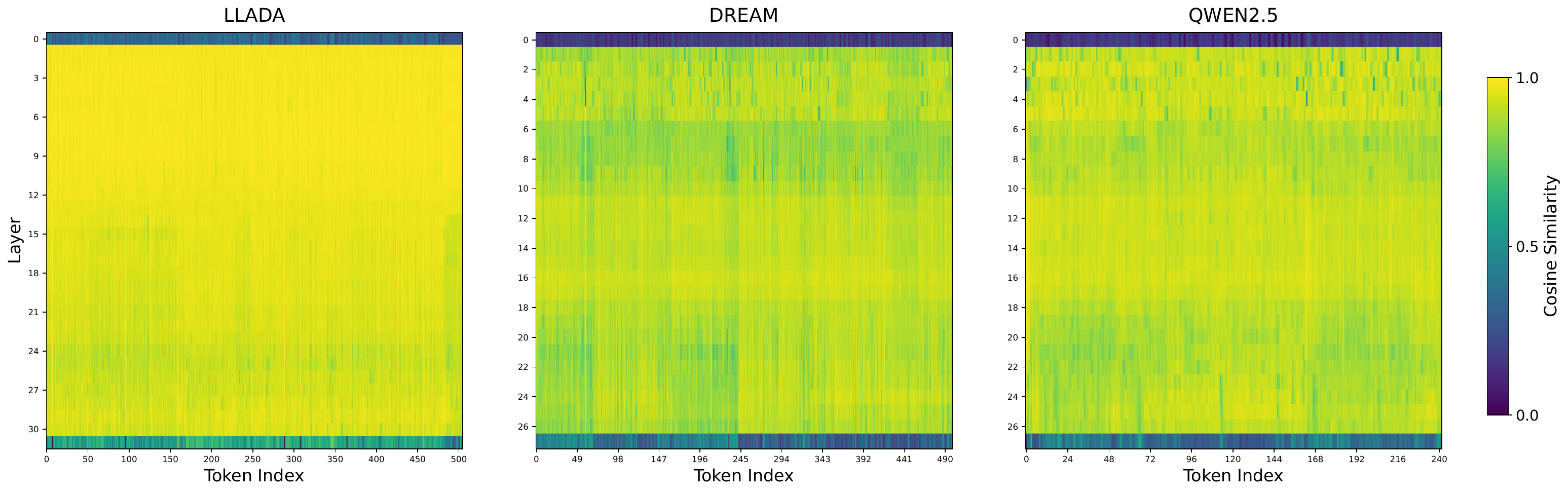}
  \caption{\textbf{Layer-wise cosine similarity across models (128 tokens decoded).} Panels show similarity between consecutive layers for (left) LLaDA, (middle) Dream-7B, and (right) Qwen2.5. High-similarity regions (yellow) indicate representational redundancy. \textbf{LLaDA exhibits a sharp two-regime structure}: layers 1 to 13 form a near uniform high similarity plateau ($>0.95$), while layers 14 to 31 show progressively more texture, with the lowest similarity concentrated in the final layers (24 to 31) where decoder-like refinement occurs. Dream-7B and Qwen2.5 lack this plateau entirely, and lower similarity bands emerge as early as layers 2 to 6 and persist through the last layer, closely tracking each other and confirming Dream-7B's strong initialization bias toward Qwen2.5's AR-like structure.}
  \label{fig:layer-sim_all}
\end{figure*}

\begin{figure*}[t]
  \centering
  \includegraphics[width=0.48\textwidth]{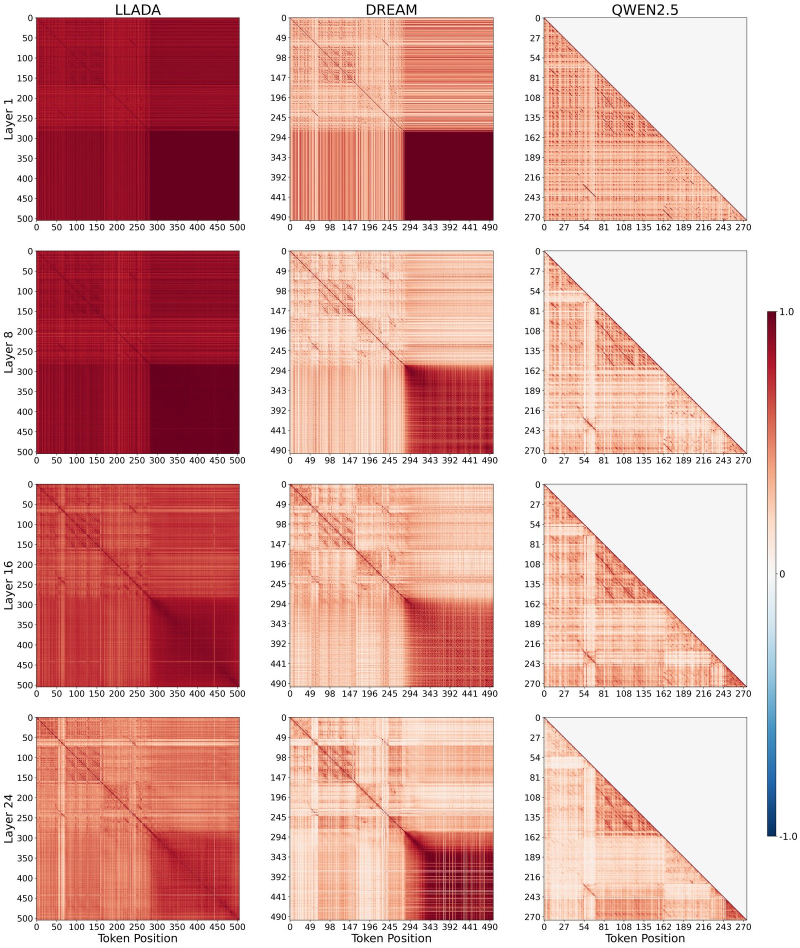}
  \hfill
  \includegraphics[width=0.48\textwidth]{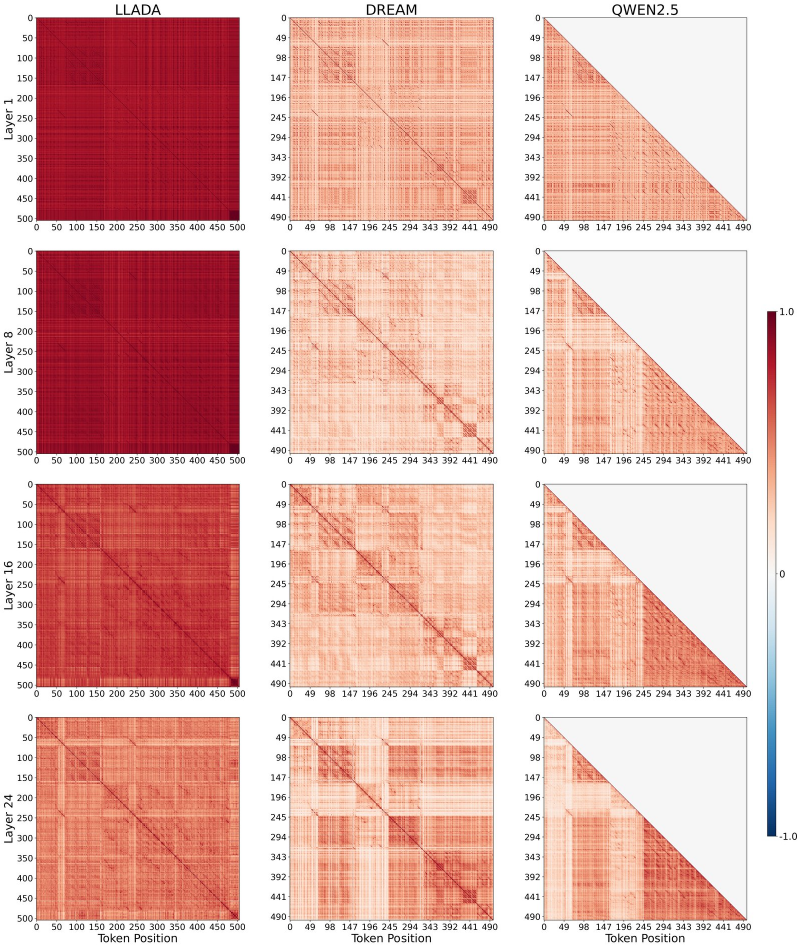}
  \caption{
    \textbf{Token-wise cosine similarity across layers and models.}
    Rows correspond to layers ($1, 8, 16, 24$); columns show (left) LLaDA,
    (middle) Dream-7B, and (right) Qwen.
    Left: decoding limited to 32 tokens highlights early representational
    stabilization in native diffusion models.
    Right: full-sequence decoding emphasizes global context integration and
    architectural differences across objectives.
  }
  \label{fig:token-sim-wide}
\end{figure*}

We hypothesize that this objective-level difference manifests as measurable representational redundancy exploitable for inference-time efficiency gains without architectural modifications or KV-cache sharing. Specifically, we analyze the rate of change of representations across both layers and tokens—examining whether diffusion training tends to produce more global representations compared to the heavily local representations of AR models—as illustrated in Fig.~\ref{fig:layer-sim_all} and Fig.~\ref{fig:token-sim-wide}.

\textbf{Methodology:} We track the cosine similarity between consecutive layer representations $\mathbf{h}_\ell$ and $\mathbf{h}_{\ell+1}$ across all tokens in a sequence. Formally, for token $i$ at layer $\ell$, we compute:
\begin{equation}
    \text{sim}(\mathbf{h}_\ell^{(i)}, \mathbf{h}_{\ell+1}^{(i)}) = \frac{\mathbf{h}_\ell^{(i)} \cdot \mathbf{h}_{\ell+1}^{(i)}}{\|\mathbf{h}_\ell^{(i)}\| \, \|\mathbf{h}_{\ell+1}^{(i)}\|}
    \label{eq:cosine_sim}
\end{equation}
and aggregate across tokens and prompts to obtain a layer-wise similarity profile.

\begin{figure}[t]
    \centering
    \includegraphics[width=\columnwidth]{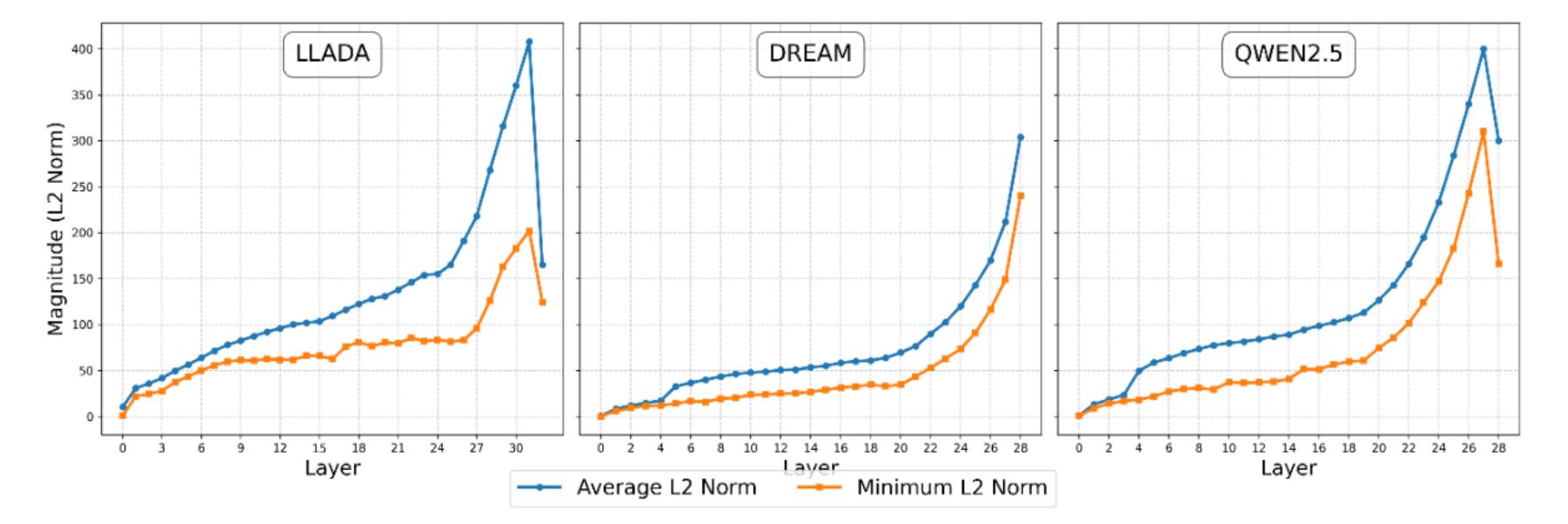}
    \caption{\textbf{Hidden-state magnitude across depth.} Layer-wise evolution of the $\ell_2$ norm of token hidden states for \textsc{LLaDA}, Dream, and Qwen. Norms remain relatively stable through the first $\sim$60--70\% of layers and increase sharply near the top of the network. The \emph{maximum} norm is dominated by rare \emph{sink tokens} (spikes; often $\ge 10^3$), so max values should be interpreted as outliers rather than typical token magnitudes.}
    \label{fig:hidden_state_norm}
\end{figure}

\textit{Why cosine similarity?} We choose cosine similarity for its robustness to magnitude changes in hidden states. Because hidden-state norms can vary substantially across layers (as we show in Figure~\ref{fig:hidden_state_norm}), a magnitude-sensitive metric would conflate representational change with scale drift. Cosine similarity isolates directional change in representation space, making it a more reliable indicator of whether a layer is performing meaningful transformation. This choice is further supported by recent work \cite{men2025shortgpt}, which demonstrates that their proposed BI metric—operating on the same mathematical rationale as cosine similarity—is the most robust among four evaluated metrics for layer-wise representational analysis. While our methodology was developed independently, this strongly reinforces our choice.

We also considered metrics like Linear Centered Kernel Alignment (CKA), a natural alternative for measuring representational similarity, but it is unsuitable for our token-wise, sample-wise analysis. CKA is designed to operate at the dataset level: it relies on mean-centering across a batch of examples, meaning a true per-token kernel cannot be defined. 
Also \cite{jiang2024tracing} empirically demonstrates that cosine similarity closely tracks CKA in layer-wise analysis. We therefore use cosine similarity directly, which is both more interpretable and computationally efficient in our per-token setting.

For dLLMs, we compute this similarity at multiple denoising steps $t \in \{1, \ldots, T\}$; for AR models, we compute it during standard forward passes. We aggregate statistics across diverse prompts from our evaluation benchmarks.

\noindent\textit{1) Global representations and early-layer redundancy in native dLLMs:} For LLaDA, from Fig.~\ref{fig:layer-sim_all} the layer-wise similarity pattern remains largely consistent across denoising steps, suggesting a tendency toward more global representational abstraction. Early layers exhibit high inter-layer similarity (plateau regions with cosine similarity $> 0.95$), while later layers and denoising steps perform iterative refinement. This organization—high redundancy in early layers, active refinement in later layers—indicates potential redundancy that can be exploited at inference time.



\noindent\textit{2) Recency bias and Global vs. Local representations:} Token-wise analysis reveals striking differences in representational dynamics. LLaDA exhibits minimal recency bias with smooth, high-similarity transitions across all tokens and layers, indicating global representational abstraction. 

In contrast, both Dream-7B and Qwen2.5 demonstrate significant recency bias—representations change substantially for each new token across all layers. Notably, in LLaDA, recency bias emerges primarily in later layers (which begin to act more like decoder layers), whereas in Dream-7B and Qwen2.5, recency bias is prominent across all layers and tokens. This suggests that AR-style models maintain consistent token-by-token representational updates throughout depth, indicating less hierarchical abstraction compared to native dLLMs.

 Our representational similarity analysis complements the behavioral analysis in \cite{gong2025diffucoder}, which measures AR-ness through generation patterns (local consecutive next-token prediction and global earliest-mask selection). While their metrics capture \textit{output-level} generation strategies—whether models follow left-to-right filling patterns—our layer-wise and token-wise cosine similarity analysis reveals \textit{internal representational dynamics}: how hidden states evolve across depth and tokens. 

 Critically, we find that Dream-7B exhibits AR-like recency bias in its \textit{representations} (mirroring Qwen2.5's token-by-token updates) despite being trained with diffusion objectives, providing mechanistic evidence for initialization bias that persists beyond surface-level generation behavior. This representational perspective explains \textit{why} certain models exhibit AR-like generation patterns and reveals that initialization effects run deeper than decoding strategies alone.

\begin{figure*}[t]
  \centering
  \includegraphics[width=0.8\textwidth]{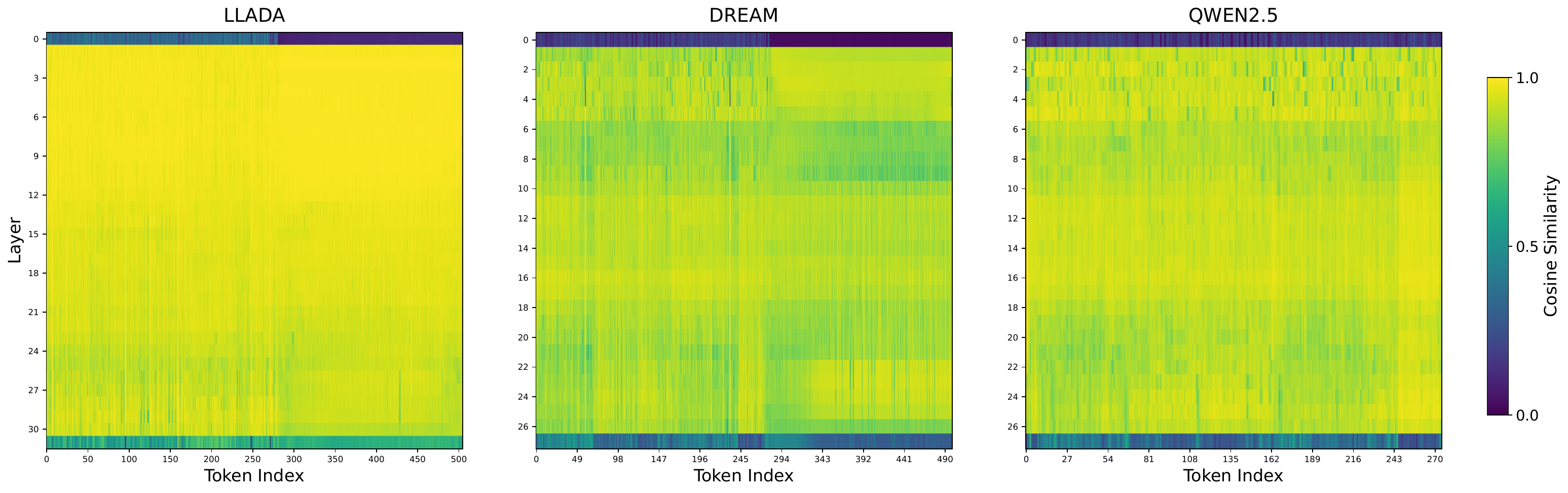}
  \caption{\textbf{Layer-wise cosine similarity across models 32 tokens decoded.} Each row shows similarity between consecutive layers for (top) LLaDA, (middle) Qwen2.5, and (bottom) Dream-7B. High-similarity regions (yellow) indicate representational redundancy. Dream-7B's pattern closely resembles Qwen2.5 despite diffusion training, revealing strong initialization bias.}
  \label{fig:layer-sim_32}
\end{figure*}

\begin{tcolorbox}[colback=black!5!white,colframe=black!75!black,
title=\textbf{Recency Bias \& Representational Abstraction}]
\textbf{Diffusion objectives reduce recency bias and promote global representations:} LLaDA shows minimal recency bias with global representations across tokens, while AR models (Qwen2.5, Dream-7B) exhibit strong recency bias at all layers. This suggests diffusion training encourages more global abstraction, whereas AR training maintains incremental, token-by-token updates throughout network depth (see Figures~\ref{fig:layer-sim_32} and~\ref{fig:token-sim-wide})
\end{tcolorbox}

\noindent\textit{3) Strong initialization bias in AR-adopted dLLMs:} Despite being trained with a diffusion objective, Dream-7B's similarity profile—both layer-wise and token-wise—closely mirrors that of its AR initialization (Qwen2.5), with high-similarity regions and recency patterns appearing in nearly identical layer ranges. 

As shown in Figure~\ref{fig:avg_token_similarity}, Dream-7B's average token-wise cosine similarity across layers follows Qwen2.5's pattern remarkably closely throughout the network depth, despite undergoing diffusion training. In contrast, LLaDA exhibits a distinctly different profile: it begins with very high similarity ($>0.95$) in early layers, indicating redundant representations with smooth transitions, then transitions to lower similarity in later layers where refinement occurs. This hierarchical pattern—high redundancy in early layers, active refinement in later layers—is characteristic of native diffusion training and absent in AR-initialized models.
This highlights the strong regularization effect of initialization on resulting representations and abstractions, persisting even after significant fine-tuning with diffusion-based objectives.

\begin{tcolorbox}[colback=black!5!white,colframe=black!75!black,
title=\textbf{Initialization Effect}]
\textbf{AR initialization creates persistent representational structure:} Dream-7B, despite diffusion training, retains Qwen2.5's similarity patterns and recency bias, demonstrating that pre-trained AR representations strongly regularize subsequent diffusion fine-tuning. Native dLLMs (LLaDA) develop fundamentally different abstraction hierarchies.
\end{tcolorbox}


\noindent\textit{Magnitude Evolution}
One potential limitation of cosine similarity is its invariance to magnitude. To ensure our redundancy findings are not artifacts of magnitude collapse, we analyze the $\ell_2$ norm of hidden states across layers as observed in Fig.~\ref{fig:hidden_state_norm}.We observed that the magnitude evolution is small for initial 60-70\% layers and then rises steeply. There is also presence of sink tokens, super high magnitude than the rest of the tokens, as discussed in \cite{rulli2025attention}.

Together, these analyses validate cosine similarity as a meaningful proxy which demonstrate potential redundancy in representations and motivate our inference-time layer-skipping strategy.

\section{Layer-skip at Inference}

The observed high-similarity plateaus suggest that certain layers contribute minimally to representational transformation. We hypothesize that \textit{skipping} these layers at inference time can reduce computational cost with minimal impact on task performance. We demonstrate with and without layer-skipping and prefix caching. 

Crucially, our approach is:1) \textit{Static and task-agnostic:} We identify skip-eligible layers based on training-time similarity analysis, without per-task tuning or dynamic routing.2) \textit{Architecture-agnostic:} Unlike YOCO-style methods that require cache-once designs or parameter sharing, our method applies to any pretrained model without modification.3) \textit{Complementary to KV-caching:} Layer skipping reduces FLOPs and depth; KV-caching reduces memory and redundant computation across tokens. Both can be combined for compounded benefits.


\textbf{Skip policy}. We define a skip set $\mathcal{L}_{\text{skip}} \subset {1,\dots,L}$ of size $N$ containing the $N$ layers whose adjacency similarity is highest, subject to the constraint that no two selected layers are consecutive. During inference, for each layer $\ell \in \mathcal{L}_{\text{skip}}$, we bypass the transformer block and directly pass $\mathbf{h}_{\ell-1}$ to layer $\ell+1$, i.e. $\mathbf{h}_\ell := \mathbf{h}_{\ell-1}$. Residual connections ensures representational continuity, so bypassing a block only discards its (small) additive update rather than truncating the residual stream. Our algorithm is shown in Algorithm~\ref{alg:static_layer_skip}.

\textbf{Quality vs. Efficiency Hypothesis:} Minimal degradation for dLLMs, High similarity indicates redundancy; skipping should preserve task performance. Larger degradation for AR models as AR models may rely more on incremental refinement, making layer skipping more disruptive.

\begin{algorithm}[h!]
\caption{Static Layer-Skip Selection}
\label{alg:static_layer_skip}
\begin{algorithmic}[1]
\Require Hidden states $\{\mathbf{h}_\ell\}_{\ell=0}^{L}$, skip budget $N$
\Ensure Skip set $\mathcal{L}_{\text{skip}}$, $|\mathcal{L}_{\text{skip}}| \le N$

\State $s_\ell \gets \cos(\mathbf{h}_{\ell-1}, \mathbf{h}_\ell)$ for $\ell = 1, \dots, L$
    \Comment{Adjacency similarity}
\State $\pi \gets \operatorname{argsort}(s_1, \dots, s_L)$ in descending order
    \Comment{Layers ranked by redundancy}
\State $\mathcal{L}_{\text{skip}} \gets \emptyset$
\For{$\ell \in \pi$}
    \If{$(\ell - 1) \notin \mathcal{L}_{\text{skip}}$ \textbf{and} $(\ell + 1) \notin \mathcal{L}_{\text{skip}}$}
        \State $\mathcal{L}_{\text{skip}} \gets \mathcal{L}_{\text{skip}} \cup \{\ell\}$
            \Comment{Greedy max-weight independent set on the layer path}
    \EndIf
    \If{$|\mathcal{L}_{\text{skip}}| = N$}
        \State \textbf{break}
    \EndIf
\EndFor
\State \Return $\mathcal{L}_{\text{skip}}$
\end{algorithmic}
\end{algorithm}



 



\section{Experimental Setup}
\label{sec:exp-setup}

\paragraph{Models.}
We evaluate three families to disentangle training objective and initialization: (i) a native diffusion LLM, \textit{LLaDA} (we use the 8B Base / Instruct checkpoints) \cite{Nie2025LLaDA}; (ii) a native autoregressive (AR) model, \textit{Qwen2.5} (7B Base / Instruct) \cite{Qwen2024Qwen25}; and (iii) an AR-initialized diffusion LLM, \textit{Dream-7B} (Instruct) \citep{Ye2025Dream7B}. Unless stated otherwise, all models are evaluated with their public inference code and default tokenizers.

\textit{Note:} Few open-source dLLMs are trained from scratch at scale; LLaDA-8B is, to our knowledge, the only such checkpoint comparable to AR baselines. Lacking a scale-matched, from-scratch alternative, we compare 7 to 8B models across families as the closest practical setting and leave fully controlled scaling comparisons to future work./
    
\paragraph{Benchmarks.}
We measure reasoning and code synthesis across standard suites: \textit{GSM8K} (grade-school math; exact-match accuracy) \citep{Cobbe2021GSM8K}; \textit{HumanEval} (function-level Python synthesis; pass@k using the official harness) \citep{Chen2021HumanEval};
\textit{MATH-500} (the 500-problem test subset of the MATH benchmark; exact-match accuracy) \citep{Hendrycks2021MATH}.

\paragraph{Prompting and answer extraction.}
For \textit{GSM8K}, we use a few-shot rationale prompt with an explicit \texttt{Final Answer:} line; we strip formatting and compare normalized numbers.
For \textit{HumanEval} and \textit{MBPP}, we request a single Python function and evaluate with the official test suites; we report pass@1 and pass@k. The same prompts are used across models to ensure comparability. We follow the exact inference setting of \cite{chen2026dflash}, including the use of \textbf{prefix caching} to mirror the Fast-dLLM evaluation setting, ensuring accurate and comparable evaluations across models.

\paragraph{Decoding \& sampling.}
For \textit{AR} decoding (Qwen2.5), we use greedy or nucleus sampling (default \texttt{top\_p}=0.95, \texttt{temperature}$\in\{0.2,0.7,0.8\}$ depending on task), \texttt{max\_new\_tokens}=2048, and early stopping on task-specific end markers. For \textit{diffusion} decoding (LLaDA, Dream-7B), we follow each repository’s default sampler/schedule and report quality–latency tradeoffs with denoising budget $T = 32$; other settings (e.g., temperature annealing or remasking) follow the public implementations \citep{Nie2025LLaDA,Ye2025Dream7B}. To compare fairly to AR decoding, we standardize context limits (2{,}048 tokens total) and stop rules.

\paragraph{Layer-skipping evaluation (ours).}
To isolate objective-induced redundancy, we introduce a \emph{static, task-agnostic} top-$k$ layer skip policy applied only at inference time, without KV sharing or architectural changes. We evaluate $k\in\{0,2,4,6\}$ on 7--8B models. Metrics include: task score (GSM8K accuracy; HumanEval/MBPP pass@1 and pass@k; MATH-500), 
and end-to-end latency/throughput (prefill$+$decode or diffusion steps).


\section{Results}
\textit{Native dLLMs Enable Aggressive Layer Skipping.}
Table~\ref{tab:model_comparison} and Figure~\ref{fig:combined} show that native diffusion LLMs are markedly more robust to layer skipping than autoregressive models. For LLaDA, skipping 6 layers (18.75\% FLOPs reduction) preserves 93--97\% of baseline performance across GSM8K, MATH500, and MBPP, with HumanEval slightly lower at 83\%. Even at an 8-layer skip (25\% FLOPs reduction), performance retention remains high (60--94\%) 
, placing LLaDA firmly in the favorable efficiency--quality regime (top right of Figure~\ref{fig:layer-skip-flops}).

In contrast, autoregressive models degrade rapidly under aggressive layer pruning, though the point of collapse differs by architecture. Qwen2.5-7B-Instruct is comparatively robust at a 2-layer skip (7.14\% FLOPs reduction), retaining 77--94\% of baseline performance across all four tasks (85\% on average)---on par with or exceeding Dream-7B-base and LLaDA at this operating point. However, this robustness collapses sharply beyond 2 layers: retention falls to 37--47\% at a 4-layer skip (14.29\% FLOPs reduction, 43\% on average) and to just 3--6\% at 6 layers. Dream-7B-base, with optimized non-consecutive layer selection, shows the opposite pattern: it trails Qwen2.5-7B-Instruct at a 2-layer skip (78\% vs.\ 85\% average) but degrades far more gracefully thereafter, retaining 50--80\% at 4 layers and 8--25\% at 6 layers---well above Qwen2.5-7B-Instruct at matched skip counts, though still below LLaDA, which stays at 87--100\% and 83--97\% at the same two points. This suggests diffusion fine-tuning partially restructures representational redundancy---enough to widen Dream-7B-base's safe operating window relative to the autoregressive baseline at higher compression---but neither retrofitted approach matches the graceful, broad-range degradation of natively trained dLLMs like LLaDA under aggressive pruning.
\textbf{Computational savings:} Native dLLMs sustain high quality retention over a wider range of FLOPs reduction than autoregressive models. LLaDA sustains 91.8\% average retention at an 18.75\% FLOPs reduction (6-layer skip), whereas Qwen2.5-7B-Instruct---despite a comparable 85\% average at a shallower 7.14\% reduction (2-layer skip)---collapses to 43\% by a 14.29\% reduction (4-layer skip). LLaDA's safe operating window thus extends roughly 2.6$\times$ further than Qwen2.5-7B-Instruct's, whose quality falls off a cliff between 2 and 4 layers skipped. These savings are orthogonal to KV-caching: layer skipping cuts depth-wise computation, while KV-caching eliminates token-wise redundancy, enabling multiplicative gains when combined.

\begin{table*}[!ht]
\renewcommand{\arraystretch}{1.2}
\caption{Performance comparison across models and layer skipping configurations.
Values represent retention percentages relative to the 0-layer baseline, with
absolute accuracy shown in parentheses for baseline rows. Retention values should be interpreted alongside absolute accuracies: large percentage swings on tasks where baseline accuracy is near floor (e.g., $\leq$0.10) reflect sampling noise rather than meaningful robustness, as small absolute fluctuations produce arbitrarily large percentage changes.}
\label{tab:model_comparison}
\resizebox{\textwidth}{!}{%
\begin{tabular}{llcccc}
\toprule
\multirow{2}{*}{Task} & \multirow{2}{*}{Layers Skipped} & \multicolumn{4}{c}{Models} \\
\cmidrule(lr){3-6}
& & \textbf{LLada-8B-Instruct} & \textbf{Dream-7B-base} & \textbf{Dream-7B-Instruct} & \textbf{Qwen2.5-7B-Instruct} \\
\midrule
\multirow{4}{*}{GSM8K} & 0 (baseline) & 100\% {\footnotesize\textcolor{gray}{(0.79)}} & 100\% {\footnotesize\textcolor{gray}{(0.74)}} & 100\% {\footnotesize\textcolor{gray}{(0.80)}} & 100\% {\footnotesize\textcolor{gray}{(0.62)}} \\
\cmidrule(lr){2-6}
& 2 & 95\% {\footnotesize\textcolor{gray}{(0.75)}} & 101\% {\footnotesize\textcolor{gray}{(0.75)}} & 91\% {\footnotesize\textcolor{gray}{(0.73)}} & 82\% {\footnotesize\textcolor{gray}{(0.51)}} \\
& 4 & 94\% {\footnotesize\textcolor{gray}{(0.74)}} & 68\% {\footnotesize\textcolor{gray}{(0.50)}} & 18\% {\footnotesize\textcolor{gray}{(0.14)}} & 42\% {\footnotesize\textcolor{gray}{(0.26)}} \\
& 6 & 94\% {\footnotesize\textcolor{gray}{(0.74)}} & 22\% {\footnotesize\textcolor{gray}{(0.16)}} & 4\% {\footnotesize\textcolor{gray}{(0.03)}} & 6\% {\footnotesize\textcolor{gray}{(0.04)}} \\
\midrule
\multirow{4}{*}{MATH500} & 0 (baseline) & 100\% {\footnotesize\textcolor{gray}{(0.38)}} & 100\% {\footnotesize\textcolor{gray}{(0.36)}} & 100\% {\footnotesize\textcolor{gray}{(0.48)}} & 100\% {\footnotesize\textcolor{gray}{(0.41)}} \\
\cmidrule(lr){2-6}
& 2 & 108\% {\footnotesize\textcolor{gray}{(0.41)}} & 69\% {\footnotesize\textcolor{gray}{(0.25)}} & 77\% {\footnotesize\textcolor{gray}{(0.37)}} & 88\% {\footnotesize\textcolor{gray}{(0.36)}} \\
& 4 & 87\% {\footnotesize\textcolor{gray}{(0.33)}} & 50\% {\footnotesize\textcolor{gray}{(0.18)}} & 31\% {\footnotesize\textcolor{gray}{(0.15)}} & 37\% {\footnotesize\textcolor{gray}{(0.15)}} \\
& 6 & 97\% {\footnotesize\textcolor{gray}{(0.37)}} & 25\% {\footnotesize\textcolor{gray}{(0.09)}} & 0\% {\footnotesize\textcolor{gray}{(0.00)}} & 5\% {\footnotesize\textcolor{gray}{(0.02)}} \\
\midrule
\multirow{4}{*}{HumanEval} & 0 (baseline) & 100\% {\footnotesize\textcolor{gray}{(0.53)}} & 100\% {\footnotesize\textcolor{gray}{(0.64)}} & 100\% {\footnotesize\textcolor{gray}{(0.62)}} & 100\% {\footnotesize\textcolor{gray}{(0.65)}} \\
\cmidrule(lr){2-6}
& 2 & 96\% {\footnotesize\textcolor{gray}{(0.51)}} & 95\% {\footnotesize\textcolor{gray}{(0.61)}} & 90\% {\footnotesize\textcolor{gray}{(0.56)}} & 94\% {\footnotesize\textcolor{gray}{(0.61)}} \\
& 4 & 91\% {\footnotesize\textcolor{gray}{(0.48)}} & 61\% {\footnotesize\textcolor{gray}{(0.39)}} & 29\% {\footnotesize\textcolor{gray}{(0.18)}} & 45\% {\footnotesize\textcolor{gray}{(0.29)}} \\
& 6 & 83\% {\footnotesize\textcolor{gray}{(0.44)}} & 8\% {\footnotesize\textcolor{gray}{(0.05)}} & 0\% {\footnotesize\textcolor{gray}{(0.00)}} & 3\% {\footnotesize\textcolor{gray}{(0.02)}} \\
\midrule
\multirow{4}{*}{MBPP} & 0 (baseline) & 100\% {\footnotesize\textcolor{gray}{(0.56)}} & 100\% {\footnotesize\textcolor{gray}{(0.44)}} & 100\% {\footnotesize\textcolor{gray}{(0.75)}} & 100\% {\footnotesize\textcolor{gray}{(0.66)}} \\
\cmidrule(lr){2-6}
& 2 & 107\% {\footnotesize\textcolor{gray}{(0.60)}} & 48\% {\footnotesize\textcolor{gray}{(0.21)}} & 80\% {\footnotesize\textcolor{gray}{(0.60)}} & 77\% {\footnotesize\textcolor{gray}{(0.51)}} \\
& 4 & 100\% {\footnotesize\textcolor{gray}{(0.56)}} & 80\% {\footnotesize\textcolor{gray}{(0.35)}} & 37\% {\footnotesize\textcolor{gray}{(0.28)}} & 47\% {\footnotesize\textcolor{gray}{(0.31)}} \\
& 6 & 93\% {\footnotesize\textcolor{gray}{(0.52)}} & 14\% {\footnotesize\textcolor{gray}{(0.06)}} & 12\% {\footnotesize\textcolor{gray}{(0.09)}} & 5\% {\footnotesize\textcolor{gray}{(0.03)}} \\
\bottomrule
\end{tabular}
}
\end{table*}

\subsection{Layer Distribution and Skip Sensitivity}

\begin{figure}
    \centering
\includegraphics[width=\linewidth]{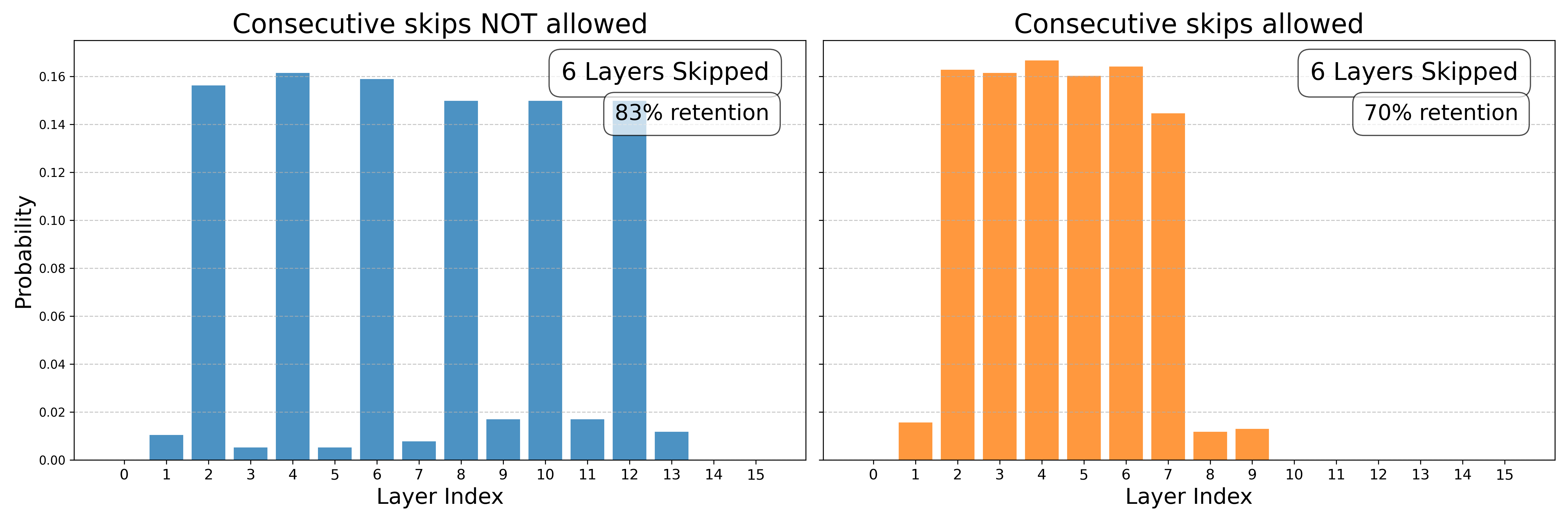}    
    \caption{\textbf{Which layers are skipped? (6-layer skip)} Distribution over layer indices on \textsc{LLaDA} (HumanEval). Disallowing consecutive skips concentrates selections in early layers (first 40--60\%).}
    \label{fig:layer_skip_dist}
    \vspace{-10pt}
\end{figure}

Table~\ref{tab:skip_sensitivity_combined} reveals that \textit{consecutive layer skipping is catastrophic}.
For LLaDA at 8-layer skip, allowing consecutive removal drops GSM8K retention from 94\% to 40\% and HumanEval from 60\% to 8\%. Similarly, accuracy drops a lot when skipping 6-8 layers uniformly or randomly justifying the utility of our layer-skip algorithm. Our algorithm
avoids this by maintaining representational continuity. Analysis and Fig.\ref{fig:layer_skip_dist} shows skipped layers concentrate in early network depth (first 40--60\%), aligning with our observation that early layers tend to develop more global representations with high redundancy, while later layers perform critical fine-grained refinement.

\begin{table}[htbp]
\centering
\renewcommand{\arraystretch}{1.1}
\caption{Comparison of Layer Skipping Strategies (Accuracy Retention) for LLaDA-8B-Instruct where Top-K is consecutive layers, Unif skips layers uniformly, and Rand skips random layers}
\label{tab:skip_sensitivity_combined}
\footnotesize
\setlength{\tabcolsep}{3pt}
\begin{tabular}{c|cccc|cccc}
\toprule
\multirow{2}{*}{Skip} & \multicolumn{4}{c|}{\textbf{GSM8K}} & \multicolumn{4}{c}{\textbf{HumanEval}} \\
\cmidrule(lr){2-5} \cmidrule(lr){6-9}
& Ours & Top-k & Unif & Rand & Ours & Top-k & Unif & Rand \\
\midrule
6 & 94\% & 95\% & 87\% & 77\% & 83\% & 70\% & 55\% & 36\% \\
8 & 94\% & 40\% & 71\% & 62\% & 60\% & 8\% & 30\% & 21\% \\
\bottomrule
\end{tabular}
\end{table}




\section{Related Work}
Diffusion language models (dLLMs) replace autoregressive decoding with bidirectional denoising objectives, enabling parallel decoding and global context modeling. Foundational work on discrete diffusion \cite{austin2021structured} led to recent dLLMs such as SEDD \cite{lou2023discrete}, and LLaDA \cite{Nie2025LLaDA,bie2025llada2}, which achieve competitive language modeling performance. Dream‑7B \cite{Ye2025Dream7B} adapts pretrained AR models to diffusion training, while MDLM \cite{sahoo2024simple} simplifies diffusion objectives. Despite this progress, how diffusion objectives shape internal representations—especially relative to AR and AR‑initialized models—remains insufficiently understood. 
Additionally, efforts have been made to integrate KV caching mechanisms to reduce redundant computation \cite{ma2025dkv, liu2025dllm}. An alternate line of work focuses on step distillation of dLLMs \cite{deschenaux2025beyond, qian2026d3llm} towards accelerating dLLM inference.

\bibliography{aaai2027}


\appendix
\section{Representational Structure and Initialization Effects.}
Prior work shows that language models organize representations hierarchically across depth, with earlier layers capturing coarse features and deeper layers refining task‑specific abstractions \cite{jawahar2019does}. While representational dynamics in AR models have been studied, systematic analyses comparing AR and diffusion models are rare. Our work provides direct, layer‑ and token‑level evidence of this initialization bias in AR‑adapted dLLMs and contrasts it with the representational redundancy that emerges in native diffusion models.

\section{Summary and Future Work}

We presented the first systematic layer- and token-wise representational analysis comparing native dLLMs, AR models, and AR-initialized dLLMs. We find that diffusion objectives produce more global representations with substantial early-layer redundancy and reduced recency bias, while AR objectives yield tightly coupled, locally-structured representations, and AR initialization imprints this structure persistently even after diffusion training. 

\textit{Deeper representational analysis:}
Cosine similarity is a first-order probe, it captures directional stability across layers but does not characterize the intrinsic information content of representations. 

\textbf{Broader model coverage and training objectives.}
Our analysis covers one native dLLM (LLaDA) and one AR-initialized dLLM (Dream-7B), leaving open whether the global representation pattern generalizes across the diffusion model family. Key open questions include: Does it extend to models with fundamentally different transition kernels, such as uniform diffusion (SEDD)? Does RL-based post-training—which has been shown to reshape representational geometry in AR models—further shift dLLMs toward or away from AR-like structure? Extending this analysis to post-trained dLLMs and multi-modal diffusion architectures would substantially broaden the scope and impact of these findings.

\section{Additional Results}
We provide accuracy retention for different numbers of layers skipped for all the models considered in experiments in \cref{tab:main_table_verbose}

\begin{table*}[htbp]
\centering
\renewcommand{\arraystretch}{1.2}
\caption{Model Performance Retention Rate Across Layer Skipping. Values represent retention percentages relative to baseline (0 layers skipped), with absolute accuracy shown in parentheses for baseline rows.}
\label{tab:main_table_verbose}
\begin{tabular}{llcccc}
\toprule
\multirow{2}{*}{Task} & \multirow{2}{*}{Layers Skipped} & \multicolumn{4}{c}{Models} \\
\cmidrule(lr){3-6}
& & \textbf{LLada-8B-Instruct} & \textbf{Dream-7B-base} & \textbf{Dream-7B-Instruct} & \textbf{Qwen2.5-7B-Instruct} \\
\midrule
\multirow{9}{*}{GSM8K} & 0 (baseline) & 100\% {\footnotesize\textcolor{gray}{(0.79)}} & 100\% {\footnotesize\textcolor{gray}{(0.74)}} & 100\% {\footnotesize\textcolor{gray}{(0.80)}} & 100\% {\footnotesize\textcolor{gray}{(0.62)}} \\
\cmidrule(lr){2-6}
& 1 & 101\% {\footnotesize\textcolor{gray}{(0.80)}} & 99\% {\footnotesize\textcolor{gray}{(0.73)}} & 99\% {\footnotesize\textcolor{gray}{(0.79)}} & 106\% {\footnotesize\textcolor{gray}{(0.66)}} \\
& 2 & 95\% {\footnotesize\textcolor{gray}{(0.75)}} & 101\% {\footnotesize\textcolor{gray}{(0.75)}} & 91\% {\footnotesize\textcolor{gray}{(0.73)}} & 82\% {\footnotesize\textcolor{gray}{(0.51)}} \\
& 3 & 100\% {\footnotesize\textcolor{gray}{(0.79)}} & 77\% {\footnotesize\textcolor{gray}{(0.57)}} & 71\% {\footnotesize\textcolor{gray}{(0.57)}} & 79\% {\footnotesize\textcolor{gray}{(0.49)}} \\
& 4 & 94\% {\footnotesize\textcolor{gray}{(0.74)}} & 68\% {\footnotesize\textcolor{gray}{(0.50)}} & 18\% {\footnotesize\textcolor{gray}{(0.14)}} & 42\% {\footnotesize\textcolor{gray}{(0.26)}} \\
& 5 & 94\% {\footnotesize\textcolor{gray}{(0.74)}} & 55\% {\footnotesize\textcolor{gray}{(0.41)}} & 4\% {\footnotesize\textcolor{gray}{(0.03)}} & 15\% {\footnotesize\textcolor{gray}{(0.09)}} \\
& 6 & 94\% {\footnotesize\textcolor{gray}{(0.74)}} & 22\% {\footnotesize\textcolor{gray}{(0.16)}} & 4\% {\footnotesize\textcolor{gray}{(0.03)}} & 6\% {\footnotesize\textcolor{gray}{(0.04)}} \\
& 7 & 104\% {\footnotesize\textcolor{gray}{(0.82)}} & 1\% {\footnotesize\textcolor{gray}{(0.01)}} & 0\% {\footnotesize\textcolor{gray}{(0.00)}} & 2\% {\footnotesize\textcolor{gray}{(0.01)}} \\
& 8 & 94\% {\footnotesize\textcolor{gray}{(0.74)}} & - & - & - \\
\midrule
\multirow{9}{*}{MATH500} & 0 (baseline) & 100\% {\footnotesize\textcolor{gray}{(0.38)}} & 100\% {\footnotesize\textcolor{gray}{(0.36)}} & 100\% {\footnotesize\textcolor{gray}{(0.48)}} & 100\% {\footnotesize\textcolor{gray}{(0.41)}} \\
\cmidrule(lr){2-6}
& 1 & 100\% {\footnotesize\textcolor{gray}{(0.38)}} & 92\% {\footnotesize\textcolor{gray}{(0.33)}} & 73\% {\footnotesize\textcolor{gray}{(0.35)}} & 90\% {\footnotesize\textcolor{gray}{(0.37)}} \\
& 2 & 108\% {\footnotesize\textcolor{gray}{(0.41)}} & 69\% {\footnotesize\textcolor{gray}{(0.25)}} & 77\% {\footnotesize\textcolor{gray}{(0.37)}} & 88\% {\footnotesize\textcolor{gray}{(0.36)}} \\
& 3 & 92\% {\footnotesize\textcolor{gray}{(0.35)}} & 53\% {\footnotesize\textcolor{gray}{(0.19)}} & 50\% {\footnotesize\textcolor{gray}{(0.24)}} & 68\% {\footnotesize\textcolor{gray}{(0.28)}} \\
& 4 & 87\% {\footnotesize\textcolor{gray}{(0.33)}} & 50\% {\footnotesize\textcolor{gray}{(0.18)}} & 31\% {\footnotesize\textcolor{gray}{(0.15)}} & 37\% {\footnotesize\textcolor{gray}{(0.15)}} \\
& 5 & 105\% {\footnotesize\textcolor{gray}{(0.40)}} & 33\% {\footnotesize\textcolor{gray}{(0.12)}} & 4\% {\footnotesize\textcolor{gray}{(0.02)}} & 12\% {\footnotesize\textcolor{gray}{(0.05)}} \\
& 6 & 97\% {\footnotesize\textcolor{gray}{(0.37)}} & 25\% {\footnotesize\textcolor{gray}{(0.09)}} & 0\% {\footnotesize\textcolor{gray}{(0.00)}} & 5\% {\footnotesize\textcolor{gray}{(0.02)}} \\
& 7 & 84\% {\footnotesize\textcolor{gray}{(0.32)}} & 0\% {\footnotesize\textcolor{gray}{(0.00)}} & 0\% {\footnotesize\textcolor{gray}{(0.00)}} & 5\% {\footnotesize\textcolor{gray}{(0.02)}} \\
& 8 & 66\% {\footnotesize\textcolor{gray}{(0.25)}} & - & - & - \\
\midrule
\multirow{9}{*}{HumanEval} & 0 (baseline) & 100\% {\footnotesize\textcolor{gray}{(0.53)}} & 100\% {\footnotesize\textcolor{gray}{(0.64)}} & 100\% {\footnotesize\textcolor{gray}{(0.62)}} & 100\% {\footnotesize\textcolor{gray}{(0.65)}} \\
\cmidrule(lr){2-6}
& 1 & 100\% {\footnotesize\textcolor{gray}{(0.53)}} & 84\% {\footnotesize\textcolor{gray}{(0.54)}} & 76\% {\footnotesize\textcolor{gray}{(0.47)}} & 92\% {\footnotesize\textcolor{gray}{(0.60)}} \\
& 2 & 96\% {\footnotesize\textcolor{gray}{(0.51)}} & 95\% {\footnotesize\textcolor{gray}{(0.61)}} & 90\% {\footnotesize\textcolor{gray}{(0.56)}} & 94\% {\footnotesize\textcolor{gray}{(0.61)}} \\
& 3 & 100\% {\footnotesize\textcolor{gray}{(0.53)}} & 66\% {\footnotesize\textcolor{gray}{(0.42)}} & 73\% {\footnotesize\textcolor{gray}{(0.45)}} & 69\% {\footnotesize\textcolor{gray}{(0.45)}} \\
& 4 & 91\% {\footnotesize\textcolor{gray}{(0.48)}} & 61\% {\footnotesize\textcolor{gray}{(0.39)}} & 29\% {\footnotesize\textcolor{gray}{(0.18)}} & 45\% {\footnotesize\textcolor{gray}{(0.29)}} \\
& 5 & 77\% {\footnotesize\textcolor{gray}{(0.41)}} & 39\% {\footnotesize\textcolor{gray}{(0.25)}} & 5\% {\footnotesize\textcolor{gray}{(0.03)}} & 18\% {\footnotesize\textcolor{gray}{(0.12)}} \\
& 6 & 83\% {\footnotesize\textcolor{gray}{(0.44)}} & 8\% {\footnotesize\textcolor{gray}{(0.05)}} & 0\% {\footnotesize\textcolor{gray}{(0.00)}} & 3\% {\footnotesize\textcolor{gray}{(0.02)}} \\
& 7 & 79\% {\footnotesize\textcolor{gray}{(0.42)}} & 0\% {\footnotesize\textcolor{gray}{(0.00)}} & 0\% {\footnotesize\textcolor{gray}{(0.00)}} & 2\% {\footnotesize\textcolor{gray}{(0.01)}} \\
& 8 & 60\% {\footnotesize\textcolor{gray}{(0.32)}} & - & - & - \\
\midrule
\multirow{9}{*}{MBPP} & 0 (baseline) & 100\% {\footnotesize\textcolor{gray}{(0.56)}} & 100\% {\footnotesize\textcolor{gray}{(0.44)}} & 100\% {\footnotesize\textcolor{gray}{(0.75)}} & 100\% {\footnotesize\textcolor{gray}{(0.66)}} \\
\cmidrule(lr){2-6}
& 1 & 107\% {\footnotesize\textcolor{gray}{(0.60)}} & 66\% {\footnotesize\textcolor{gray}{(0.29)}} & 96\% {\footnotesize\textcolor{gray}{(0.72)}} & 97\% {\footnotesize\textcolor{gray}{(0.64)}} \\
& 2 & 107\% {\footnotesize\textcolor{gray}{(0.60)}} & 48\% {\footnotesize\textcolor{gray}{(0.21)}} & 80\% {\footnotesize\textcolor{gray}{(0.60)}} & 77\% {\footnotesize\textcolor{gray}{(0.51)}} \\
& 3 & 97\% {\footnotesize\textcolor{gray}{(0.54)}} & 91\% {\footnotesize\textcolor{gray}{(0.40)}} & 68\% {\footnotesize\textcolor{gray}{(0.51)}} & 65\% {\footnotesize\textcolor{gray}{(0.43)}} \\
& 4 & 100\% {\footnotesize\textcolor{gray}{(0.56)}} & 80\% {\footnotesize\textcolor{gray}{(0.35)}} & 37\% {\footnotesize\textcolor{gray}{(0.28)}} & 47\% {\footnotesize\textcolor{gray}{(0.31)}} \\
& 5 & 97\% {\footnotesize\textcolor{gray}{(0.54)}} & 32\% {\footnotesize\textcolor{gray}{(0.14)}} & 11\% {\footnotesize\textcolor{gray}{(0.08)}} & 17\% {\footnotesize\textcolor{gray}{(0.11)}} \\
& 6 & 93\% {\footnotesize\textcolor{gray}{(0.52)}} & 14\% {\footnotesize\textcolor{gray}{(0.06)}} & 12\% {\footnotesize\textcolor{gray}{(0.09)}} & 5\% {\footnotesize\textcolor{gray}{(0.03)}} \\
& 7 & 89\% {\footnotesize\textcolor{gray}{(0.50)}} & 3\% {\footnotesize\textcolor{gray}{(0.01)}} & 0\% {\footnotesize\textcolor{gray}{(0.00)}} & 0\% {\footnotesize\textcolor{gray}{(0.00)}} \\
& 8 & 77\% {\footnotesize\textcolor{gray}{(0.43)}} & - & - & - \\
\bottomrule
\end{tabular}
\end{table*}

We show wall-time savings for LLada-8B-Instruct model is \cref{tab:llada_performance_savings} showing that our method results in faster inference with minimal accuracy drop. Furthermore, we integrate our method with dual-cache in \cref{tab:gsm8k_dual_cache} and show that accuracy retention is still 88\% when skipping 4 layers, that is, 12.5\% FLOPs saving
\begin{table*}[htbp]
\centering
\renewcommand{\arraystretch}{1.2}
\caption{\textbf{LLada-8B-Instruct} Performance and Computation Savings (wall-time reduction) on various downstream tasks}
\label{tab:llada_performance_savings}
\begin{tabular}{l|c|ccccc}
\toprule
\multirow{2}{*}{Task} & \multirow{2}{*}{Metric} & \multicolumn{4}{c}{Layers Skipped} \\
\cmidrule(lr){3-7}
& & 0 (baseline) & 2 & 4 & 6 & 8 \\
\midrule
\multirow{2}{*}{\textbf{GSM8K}} & Accuracy retention & 100\% & 95\% & 94\% & 94\% & 94\% \\
& Wall time reduction & 0\% & -8\% & 2\% & 4\% & 10\% \\
\midrule
\multirow{2}{*}{\textbf{MATH500}} & Accuracy retention & 100\% & 108\% & 87\% & 97\% & 66\% \\
& Wall time reduction & 0\% & -3\% & 2\% & 10\% & 29\% \\
\midrule
\multirow{2}{*}{\textbf{HumanEval}} & Accuracy retention & 100\% & 96\% & 91\% & 83\% & 60\% \\
& Wall time reduction & 0\% & 0\% & 17\% & 21\% & 27\% \\
\midrule
\multirow{2}{*}{\textbf{MBPP}} & Accuracy retention & 100\% & 107\% & 100\% & 93\% & 77\% \\
& Wall time reduction & 0\% & 12\% & 16\% & 25\% & 29\% \\
\bottomrule
\end{tabular}
\end{table*}
\begin{table*}[htbp]
\centering
\renewcommand{\arraystretch}{1.1}
\caption{\textbf{Llada-8B-Instruct} evaluated on GSM8K with Layer Skipping and Dual Cache Method}
\label{tab:gsm8k_dual_cache}
\begin{tabular}{cc}
\toprule
\textbf{Layers Skipped} & \textbf{Retention Score} \\
\midrule
0 (baseline) & 100 {\footnotesize\textcolor{gray}{(0.78)}} \\
2 & 104 {\footnotesize\textcolor{gray}{(0.81)}} \\
3 & 101 {\footnotesize\textcolor{gray}{(0.79)}} \\
4 & 88 {\footnotesize\textcolor{gray}{(0.69)}} \\
5 & 77 {\footnotesize\textcolor{gray}{(0.60)}} \\
6 & 68 {\footnotesize\textcolor{gray}{(0.53)}} \\
\bottomrule
\end{tabular}
\end{table*}

\section{Additional Analysis and Visualizations}



\subsection{Detailed Token-wise Similarity Analysis}

We provide comprehensive token-wise similarity visualizations to complement the layer-wise analysis presented earlier. These reveal how hidden state representations evolve across tokens within individual layers, providing deeper insight into the recency bias and global vs. local representation patterns discussed in the main text.

\begin{figure*}[t]
  \centering
  \includegraphics[width=0.95\textwidth,trim=0 120pt 20 70pt, clip]{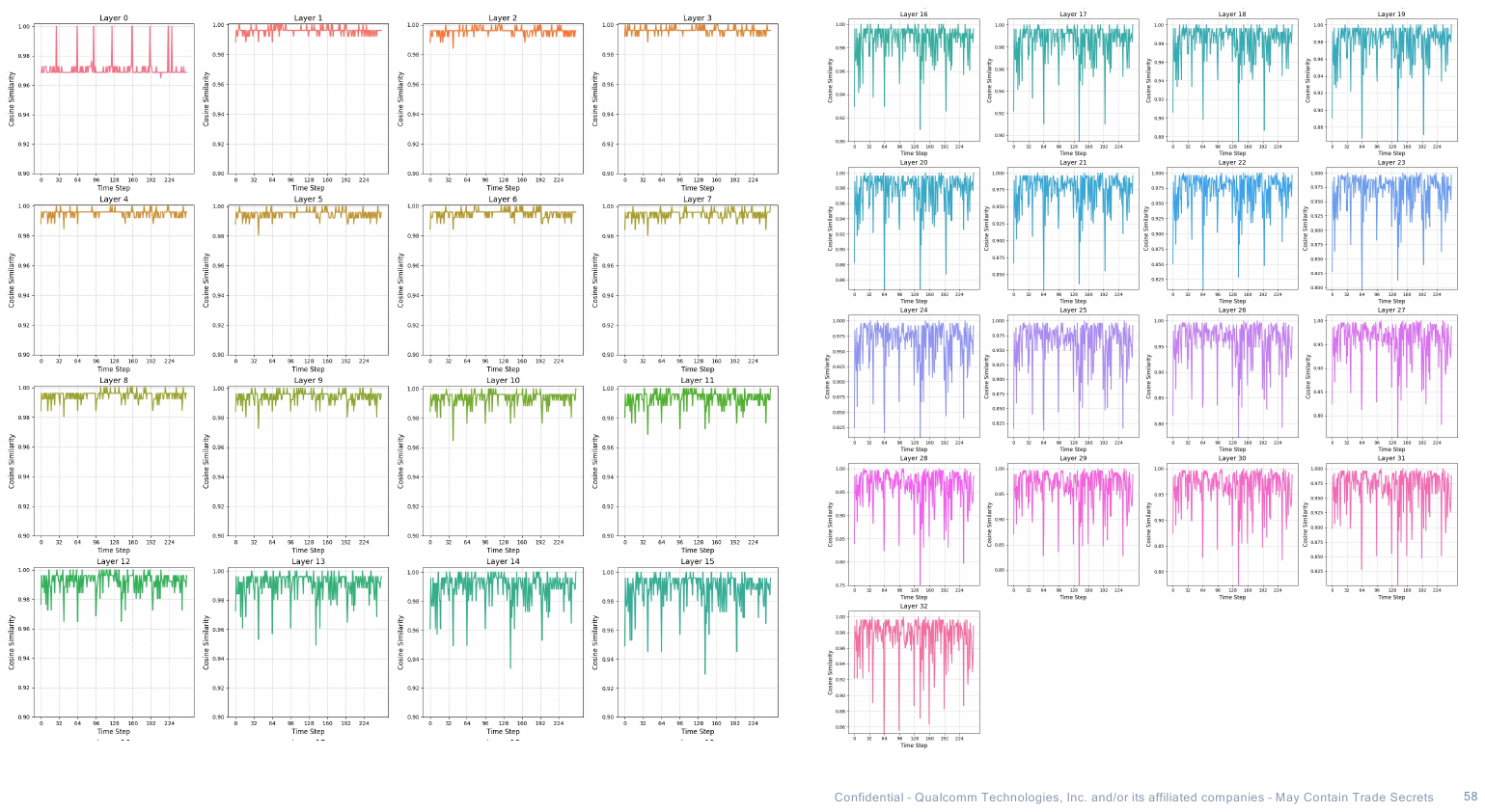}
  \caption{\textbf{Token-wise cosine similarity across all layers for LLaDA.} Each subplot shows the cosine similarity between consecutive token representations ($\mathbf{h}_{\ell,i}$ and $\mathbf{h}_{\ell,i+1}$) within a specific layer $\ell$. High similarity indicates smooth representational transitions, while low similarity indicates significant representational changes between tokens. LLaDA exhibits consistently high token-wise similarity across early layers, demonstrating minimal recency bias and global representational abstraction. Later layers show increased variability, indicating task-specific refinement and decoder-like behavior. This pattern validates our hypothesis that native dLLMs develop coarse-to-fine abstraction hierarchies, with early layers establishing stable global context and later layers performing iterative refinement.}
  \label{fig:llada_tokenwise_similarity}
\end{figure*}

Figure~\ref{fig:llada_tokenwise_similarity} shows token-wise cosine similarity across all 32 layers of LLaDA. Early layers (0--15) exhibit consistently high similarity ($>0.9$) between consecutive tokens, indicating smooth representational transitions with minimal recency bias. This validates our hypothesis that native dLLMs establish stable global context in early layers. Later layers (16--31) show increased variability and lower similarity, reflecting task-specific refinement and decoder-like behavior where representations are actively updated for generation.

\begin{figure*}[t]
  \centering
  \includegraphics[width=0.95\textwidth,trim=0 120pt 20 70pt, clip]{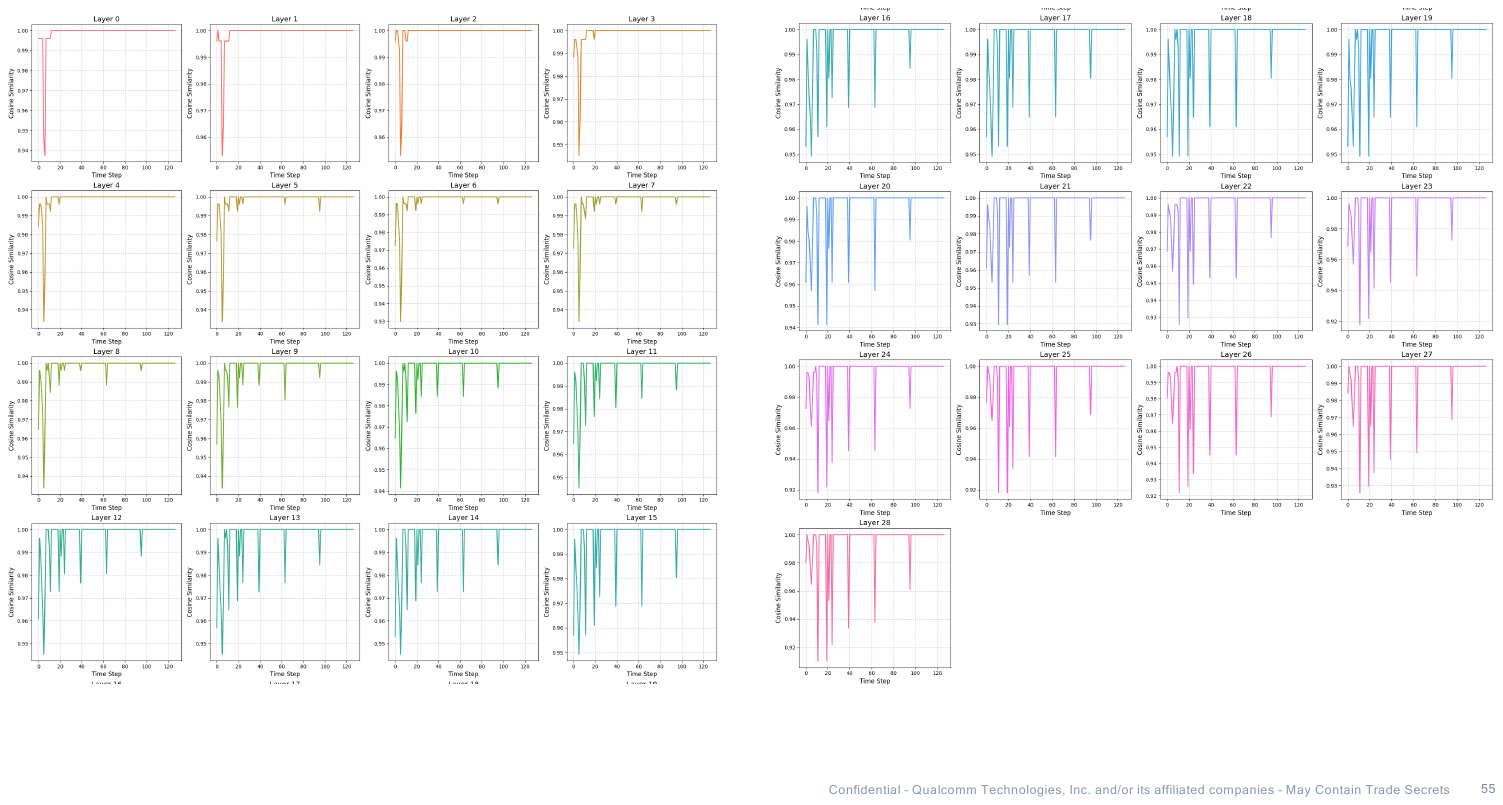}
  \caption{\textbf{Token-wise cosine similarity across all layers for Dream-7B.} Each subplot shows the cosine similarity between consecutive token representations within a specific layer. In stark contrast to LLaDA (Figure~\ref{fig:llada_tokenwise_similarity}), Dream-7B exhibits significant recency bias across \textit{all} layers, with substantial representational changes for each new token often. 
  The lack of hierarchical abstraction—with similar token-by-token update patterns across all depths—confirms that Dream-7B retains AR-like incremental refinement and retain different representational abstraction compared to native dLLMs. This provides mechanistic evidence for the initialization bias observed in our layer-skip experiments (Table~\ref{tab:model_comparison}), where Dream-7B's brittleness mirrors Qwen2.5 despite diffusion training.}
  \label{fig:dream_tokenwise_similarity}
\end{figure*}

In stark contrast, from Fig. ~\ref{fig:dream_tokenwise_similarity} reveals that Dream-7B maintains significant recency bias across \textit{all} layers. Consecutive token representations show substantial changes throughout network depth, mirroring the incremental token-by-token refinement characteristic of autoregressive models. This pattern persists despite diffusion training, providing mechanistic evidence that AR initialization creates persistent representational structure. The lack of hierarchical abstraction—with similar update patterns across all depths—explains Dream-7B's brittleness under layer skipping (Table~\ref{tab:model_comparison}), where it behaves more like Qwen2.5 than LLaDA.

\subsection{Layer-wise Token Similarity by Depth}

Figures below show token-wise similarity patterns grouped by network depth, revealing the transition from global to local representations:

\begin{figure}[t]
    \centering
    \begin{subfigure}[t]{0.48\linewidth}
        \centering
        \includegraphics[width=\linewidth]{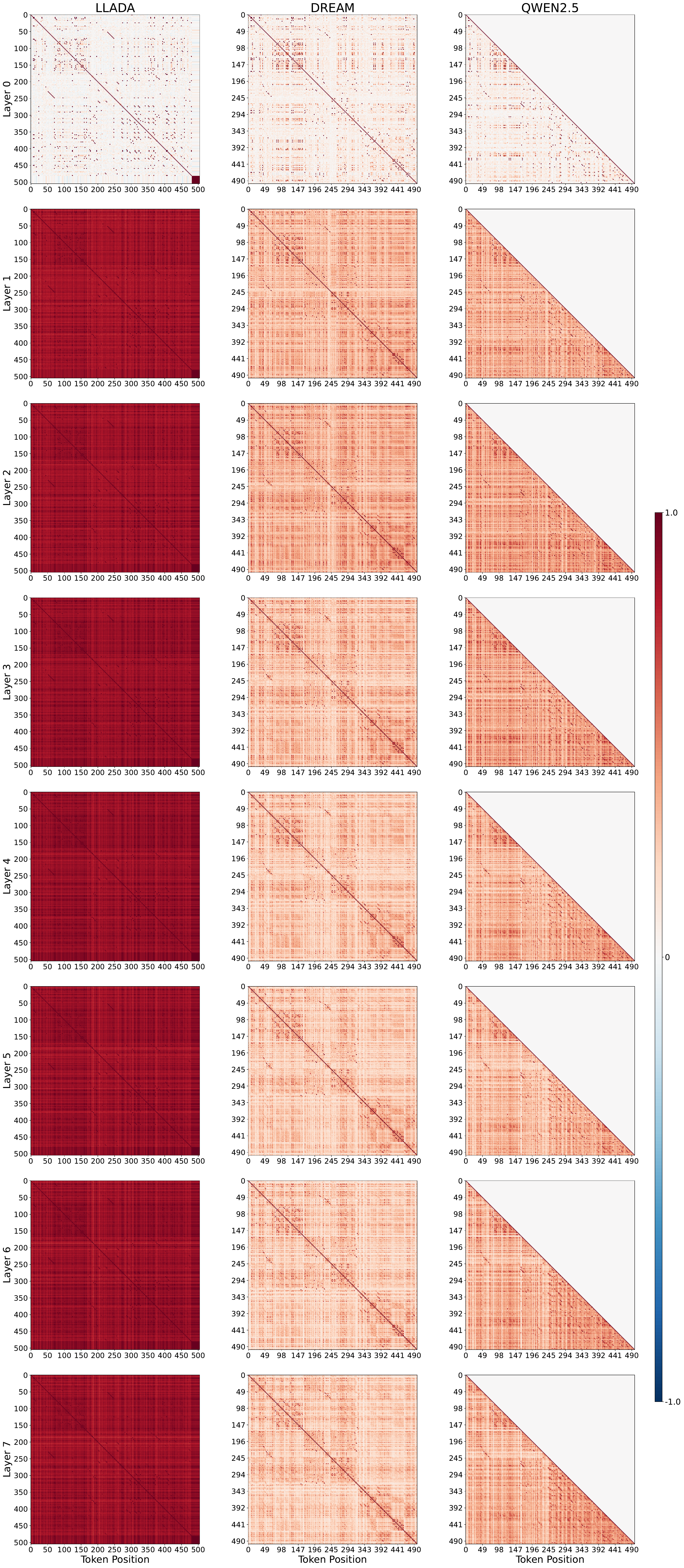}
        \caption{\textbf{Token-wise similarity in early layers (0--7).} LLaDA shows uniformly high similarity across all tokens, indicating stable global representations. Dream-7B and Qwen2.5 exhibit lower similarity with visible recency effects, demonstrating incremental AR-style processing even in early layers.}
        \label{fig:token_similarity_07}
    \end{subfigure}
    \hfill
    \begin{subfigure}[t]{0.48\linewidth}
        \centering
        \includegraphics[width=\linewidth]{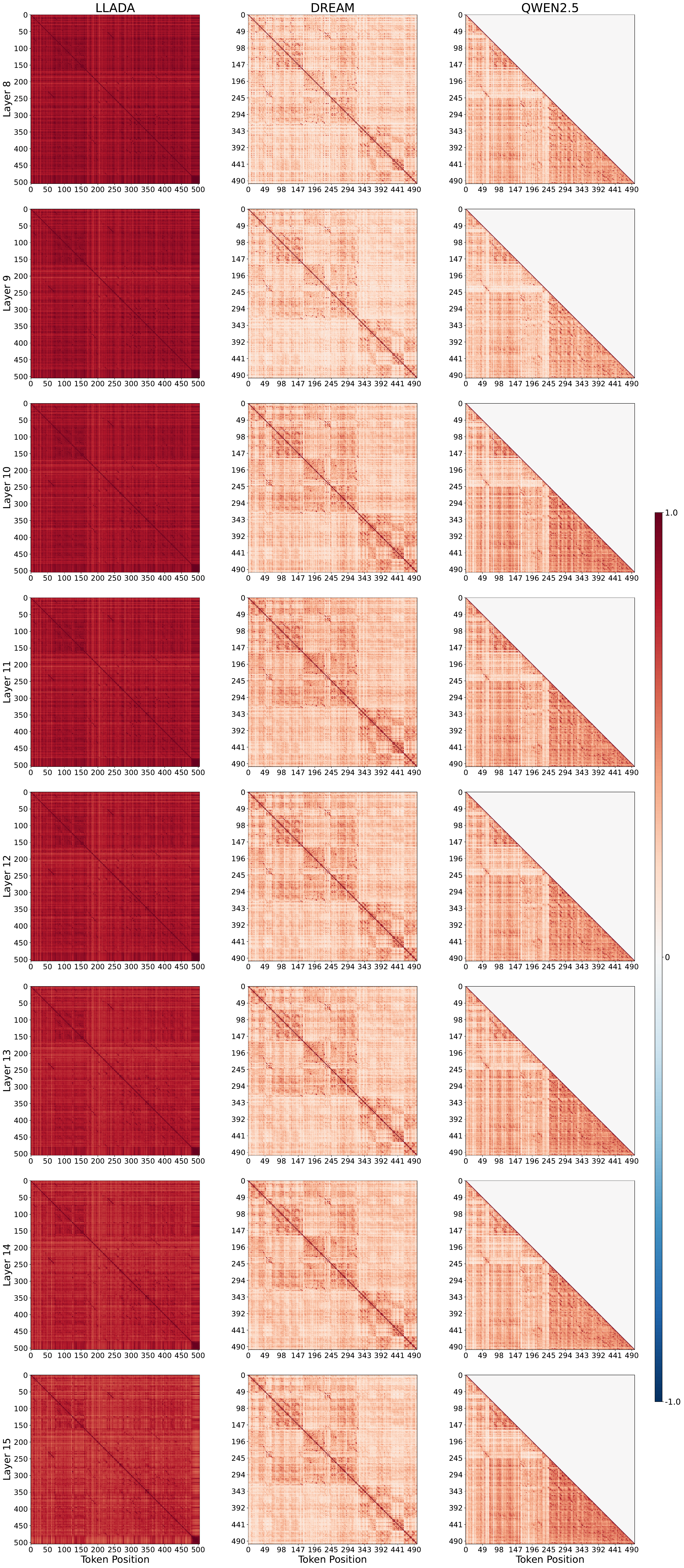}
        \caption{\textbf{Token-wise similarity in early-middle layers (8--15).} LLaDA maintains high similarity, while Dream-7B and Qwen2.5 continue showing strong recency bias. The divergence between native dLLM and AR-initialized models becomes more pronounced.}
        \label{fig:token_similarity_15}
    \end{subfigure}
    \caption{\textbf{Token-wise similarity across early and early-middle layers.} (a) Layers 0--7 and (b) layers 8--15.}
    \label{fig:token_similarity_side_by_side}
\end{figure}

\begin{figure}[t]
    \centering
    \begin{subfigure}[t]{0.48\linewidth}
        \centering
        \includegraphics[width=\linewidth]{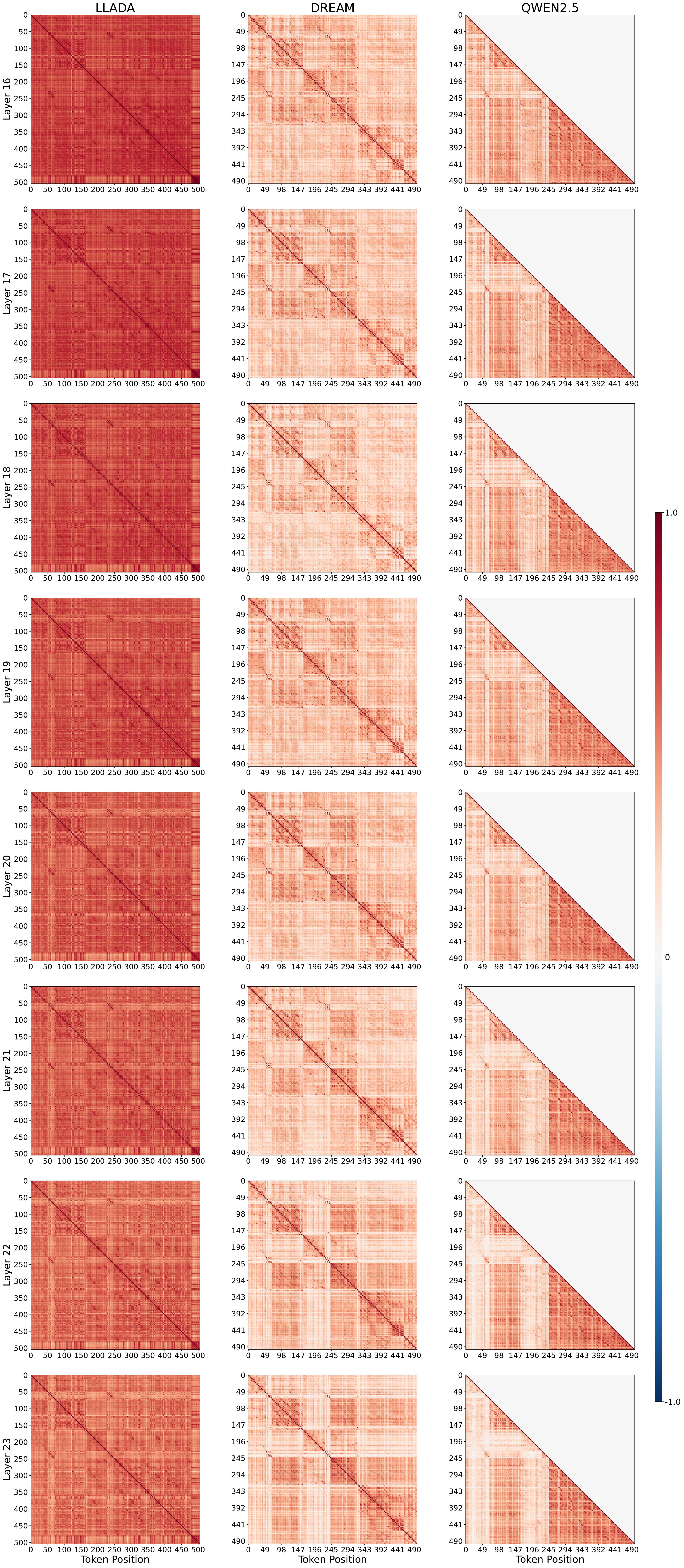}
        \caption{\textbf{Late-middle layers (16--23).} LLaDA begins transitioning to lower similarity, indicating the onset of task-specific refinement. Dream-7B and Qwen2.5 maintain consistent recency patterns throughout depth.}
        \label{fig:token_similarity_23}
    \end{subfigure}
    \hfill
    \begin{subfigure}[t]{0.48\linewidth}
        \centering
        \includegraphics[width=\linewidth]{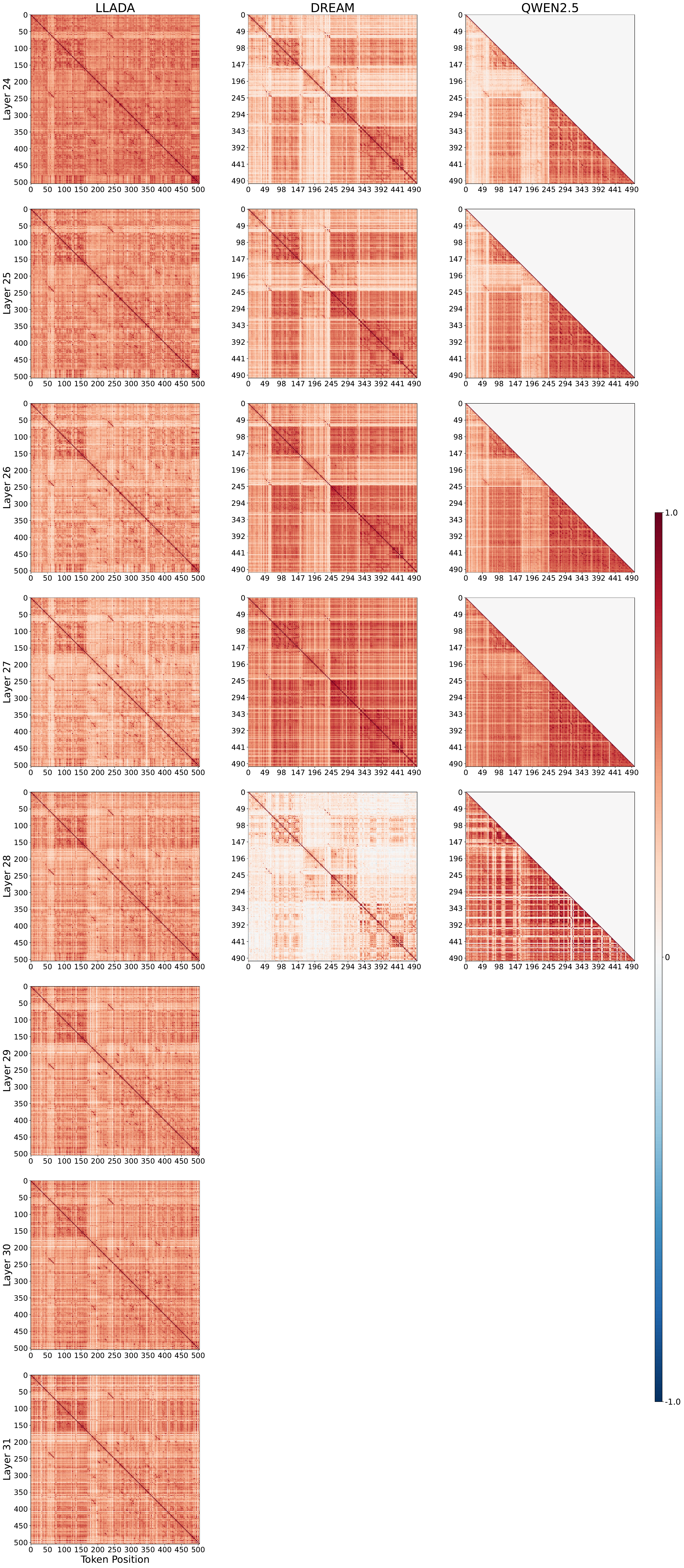}
        \caption{\textbf{Late layers (24--31).} LLaDA shows increased variability and lower similarity, reflecting active decoder-like refinement for generation. Dream-7B and Qwen2.5 continue incremental updates with strong recency bias, lacking the global transition observed in native dLLMs.}
        \label{fig:token_similarity_31}
    \end{subfigure}
    \caption{\textbf{Token-wise similarity in later layers.} (a) Late-middle (16--23) and (b) late (24--31).}
    \label{fig:token_similarity_late_side_by_side}
\end{figure}

\end{document}